\documentclass[sn-mathphys,Numbered]{sn-jnl}

\usepackage{graphicx}%
\usepackage{multirow}%
\usepackage{amsmath,amssymb,amsfonts}%
\usepackage{amsthm}%
\usepackage{mathrsfs}%
\usepackage[title]{appendix}%
\usepackage{xcolor}%
\usepackage{textcomp}%
\usepackage{manyfoot}%
\usepackage{booktabs}%
\usepackage{algorithm}%
\usepackage{algorithmicx}%
\usepackage{algpseudocode}%
\usepackage{listings}%
\usepackage{orcidlink}%
\usepackage{svg}%
\usepackage{siunitx}%
\usepackage{pgfplots}%
\usepackage{acronym}%

\pgfplotsset{compat=1.18}

\begin{document}

\acrodef{LExCI}{Learning and Experiencing Cycle Interface}
\acrodef{AI}{artificial intelligence}
\acrodef{ML}{machine learning}
\acrodef{RL}{reinforcement learning}
\acrodef{TL}{transfer learning}
\acrodef{MDP}{Markov decision process}
\acrodef{NN}{neural network}
\acrodef{ANN}{artificial neural network}
\acrodef{RCP}{rapid control prototyping}
\acrodef{PPO}{Proximal Policy Optimization}
\acrodef{DDPG}{Deep Deterministic Policy Gradient}
\acrodef{DQN}{Deep Q-Network}
\acrodef{SGD}{stochastic gradient descent}
\acrodef{KL}{Kullback-Leibler}
\acrodef{OU}{Ornstein–Uhlenbeck}
\acrodef{ADAS}{advanced driver-assistance systems}
\acrodef{IT}{information technology}
\acrodef{CPU}{central processing unit}
\acrodef{OS}{operating system}
\acrodef{PC}{personal computer}
\acrodef{ONNX}{Open Neural Network Exchange}
\acrodef{TF}{TensorFlow}
\acrodef{TF Lite}{TensorFlow Lite}
\acrodef{TFLM}{TensorFlow Lite Micro}
\acrodef{RNN}{recurrent neural network}
\acrodef{LSTM}{long short-term memory}
\acrodef{CNN}{convolutional neural network}
\acrodef{API}{application programming interface}
\acrodef{XiL}{X-in-the-loop}
\acrodef{HiL}{hardware-in-the-loop}
\acrodef{EGR}{exhaust gas recirculation}
\acrodef{ECU}{engine control unit}
\acrodef{LR}{learning rate}
\acrodef{MABX}{MicroAutoBox}
\acrodef{VF}{value function}
\acrodef{IT}{information technology}
\acrodef{VGT}{variable-geometry turbocharger}

\title[Article Title]{LExCI: A Framework for Reinforcement Learning with Embedded Systems\footnote{\tiny{This preprint has not undergone peer review or any post-submission improvements or corrections. The Version of Record of this article is published in Applied Intelligence, and is available online at \url{https://doi.org/10.1007/s10489-024-05573-0}.}}}


\author[1]{\fnm{Kevin} \sur{Badalian}\orcidlink{0000-0002-5593-0227}}\email{badalian\_k@mmp.rwth-aachen.de}

\author[1]{\fnm{Lucas} \sur{Koch}\orcidlink{0000-0002-7368-8833}}\email{koch\_luc@mmp.rwth-aachen.de}

\author[1]{\fnm{Tobias} \sur{Brinkmann}\orcidlink{0000-0003-2165-0371}}\email{brinkmann@mmp.rwth-aachen.de}

\author[1]{\fnm{Mario} \sur{Picerno}\orcidlink{0000-0002-8835-3040}}\email{picerno\_ma@mmp.rwth-aachen.de}

\author[2]{\fnm{Marius} \sur{Wegener}\orcidlink{0000-0001-8715-8609}}\email{wegener@fev.com}

\author[1]{\fnm{Sung-Yong} \sur{Lee}\orcidlink{0000-0002-9246-8895}}\email{lee\_sun@mmp.rwth-aachen.de}

\author*[1]{\fnm{Jakob} \sur{Andert}\orcidlink{0000-0002-6754-1907}}\email{andert@mmp.rwth-aachen.de}

\affil[1]{\orgname{Teaching and Research Area Mechatronics in Mobile Propulsion, RWTH Aachen University}, \orgaddress{\street{Forckenbeckstraße 4}, \city{Aachen}, \postcode{52074}, \state{NRW}, \country{Germany}}}

\affil[2]{\orgname{FEV Europe GmbH}, \orgaddress{\street{Neuenhofstraße 181}, \city{Aachen}, \postcode{52078}, \state{NRW}, \country{Germany}}}


\abstract{Advances in artificial intelligence (AI) have led to its application in many 
areas of everyday life.
In the context of control engineering, reinforcement learning (RL) represents a 
particularly promising approach as it is centred around the idea of allowing an agent to freely 
interact with its environment to find an optimal strategy. One of the challenges professionals face when 
training and deploying RL agents is that the latter often have to run on dedicated 
embedded devices. This could be to integrate them into an existing toolchain or 
to satisfy certain performance criteria like real-time constraints. 
Conventional RL libraries, however, cannot be easily utilised in conjunction with 
that kind of hardware. In this paper, we present a framework named LExCI, the \emph{Learning and 
Experiencing Cycle Interface}, which bridges this gap and provides end-users with a free and open-source tool for 
training agents on embedded systems using the open-source library RLlib. Its operability
is demonstrated with two state-of-the-art RL-algorithms and a rapid control prototyping system.}




\maketitle

\section{Introduction}
\label{sec-introduction}

\subsection{RL, Control Tasks, and Embedded Systems}
\label{subsec:rl-control-tasks-and-embedded-systems}
In recent years, \ac{AI} has evolved into a 
scientific discipline with tangible effects on the 
lives of ordinary people. Not only 
does it allow for convenience features such as speech 
recognition or auto-completion when writing \cite{howard2019artificial}, but it is also 
increasingly being utilised to control complex devices and even safety-critical systems \cite{9237327, eurostat-ai-in-enterprises}. Modern \ac{ADAS}, not 
to mention autonomous driving, would be unimaginable 
without it \cite{Grigorescu_2020}.

\Ac{RL} is an especially useful area of 
\ac{AI} when it comes to control tasks. Since it is 
based on agents that learn through their own interactions
with the environment (i.e. they generate their own training data), \ac{RL} 
has the potential to find optimal solutions to
non-trivial problems with minimal input from experts.
One problem engineers have to address, though, is the
integration of the RL agent into the system
it shall control.
Industrial applications often come with a long list of strict requirements regarding their \ac{IT} 
ecosystems: physical space, power, and cooling capacity are usually limited \cite{electronics8111289}. At the 
same time, devices need to be rugged enough to withstand vibrations or extreme fluctuations in 
temperature. Beyond such hardware-related matters, a great number of use-cases necessitate a real-time \ac{OS} 
which guarantees that computations are performed within a fixed time window \cite{electronics8111289}. Then, there is the cost factor. High-performance components needlessly drive up the prices of commercial products if their potential is not fully harnessed. A cheaper device is therefore more favourable so long as it is adequate for its task \cite{FoundationsOfEmbeddedSystems}.

As a consequence of these boundary conditions, traditional \acp{PC} are not suitable for a wide range of applications. 
Professionals choose embedded systems instead: dedicated computers that are integrated into a larger system for the purpose of controlling the same \cite{FoundationsOfEmbeddedSystems}. Embedded systems are designed from the ground up to meet the requirements outlined above.
Nonetheless, they can be incapable of running programs 
intended for conventional computers due to their inherent limitations. Established \ac{RL} libraries like 
Ray/RLlib\footnote{\url{https://docs.ray.io/en/latest/rllib/index.html}} \cite{moritz2018ray, liang2018rllib} or Stable-Baselines3\footnote{\url{https://stable-baselines3.readthedocs.io/en/master/}} \cite{JMLR:v22:20-1364} further 
rely on third-party software (e.g. Python) which might take up too much data storage space or simply not be available on the target platform/\ac{OS}. Part of that list of dependencies
are libraries for \ac{ML} models, i.e. the mathematical structures (most notably \acp{NN}) which, among other things,
represent the behaviour of the agent. Prominent exemplars --- for instance \ac{TF}\footnote{\url{https://www.tensorflow.org/}}
\cite{tensorflow2015-whitepaper} or PyTorch\footnote{\url{https://pytorch.org/}} \cite{Paszke_PyTorch_An_Imperative_2019} ---
suffer from the same problems, meaning that merely executing a trained agent on an embedded system may not be a straightforward endeavour \cite{10.1145/3398209}.

\subsection{Model Execution}
\label{subsec:model-execution}
Even if a \ac{ML} library cannot be installed on an embedded device, there are still ways to put its 
agents to use. The simplest is to run them on external machines that are then contacted by embedded devices in order to retrieve actions for their observations.
Due to the latency associated with this option, it is likely to be
sub-optimal. Another detracting factor is that the agents are not
executed \emph{on} the actual controllers. A more fitting solution is to convert the models to a format that is suitable for the target 
hardware, possibly by translating them into a generic representation like
\ac{ONNX}\footnote{\url{https://onnx.ai/}} or some other intermediate format first.

\Ac{TFLM}\footnote{\url{https://www.tensorflow.org/lite/microcontrollers}} \cite{DBLP:journals/corr/abs-2010-08678}, for example, condenses \ac{TF} to its core functionality, optimises its code for 
micro-controllers \cite{10.1145/3398209}, and reduces the number of third-party dependencies. \Ac{TFLM} can be thought of as a subset of \ac{TF Lite}\footnote{\url{https://www.tensorflow.org/lite}}, a lean version of the full library geared towards mobile and edge devices. It is hence capable of reading \ac{TF Lite} models as long as they are  comprised of common operations. Conveniently, \ac{TF} can natively convert full models to \ac{TF Lite}.

cONNXr \cite{cONNXr}, on the other hand, is agnostic to the model's original framework as it is written to work with models defined in the \ac{ONNX} format. Other libraries take a more puristic approach and implement their own model formats in C or C++ using either nothing 
but the respective standard library or just a handful of header-only libraries. Projects in that category are Genann \cite{Genann}, KANN \cite{KANN}, tiny-dnn \cite{tiny-dnn}, or 
MiniDNN \cite{MiniDNN}. End-users have to manually re-write and configure their models with those solutions, though, because they typically lack converters.

Besides the above, there are solutions that transpile existing model formats to pure C/C++ code which is then 
compiled for the target hardware \cite{electronics8111289}. frugally-deep \cite{frugally-deep}, keras2cpp \cite{keras2cpp}, or 
onnx2c \cite{onnx2c} follow that philosophy.
Likewise, MATLAB\footnote{\url{https://www.mathworks.com/products/matlab.html}} is capable of generating code from imported \ac{ONNX} models when using its Reinforcement Learning Toolbox \cite{matlabrltoolbox}.

\subsection{Training the Model}
\label{subsec:training-the-model}
Training --- that is the act of updating an agent's model --- is performed using \ac{RL} libraries like the 
aforementioned Ray/RLlib. Given the limitations of most embedded systems, this
step is usually outsourced to a powerful workstation or a cluster so as to merely deploy the agent on the target hardware \cite{electronics8111289}. If not automated, this TinyML \cite{han2022tinyml}
strategy becomes tedious when learning with on-policy \ac{RL} algorithms
(cf. Sec. \ref{subsec:reinforcement-learning}) due to the fact that the deployment process must be repeated after
each and every modification of the agent. To make matters worse, the generated training data usually cannot be passed
directly to the algorithm either and requires post-processing.

\subsection{Proposed Solution}
\label{subsec:proposed-solution}
Motivated by the shortcomings of \ac{RL} software in this area,
we developed the \emph{Learning and Experiencing Cycle Interface} or LExCI for short.
This general-purpose framework allows experts to easily train \ac{RL} agents with Ray/RLlib when model execution happens on an embedded system and training takes place on another, conventional machine. All models are implemented in \ac{TFLM}/\ac{TF Lite} and \ac{TF}, respectively. LExCI is open-source and freely available to the public through its official GitHub 
repository (see Sec. \ref{subsec:code-availability} for the link). Our contributions are:
\begin{itemize}
  \item a free, open-source, general-purpose \ac{RL} framework based on established libraries
  \item training with embedded systems
  \item out-of-the-box support for elaborate \ac{NN} architectures such as \acp{RNN} and \acp{CNN}
  \item compatibility with different model-free \ac{RL} algorithms, both on- and off-policy (see Sec. \ref{subsec:reinforcement-learning})
  \item helper classes for automating various pieces of control software
  \item an architecture that lends itself to parallelisation
\end{itemize}

Earlier versions of the software have already proven themselves in 
academic research. In \cite{KOCH2023105477} and 
\cite{picerno2023transfer}, an agent was trained to control the high-
pressure \ac{EGR} valve of a Euro 6d Diesel engine on different 
\ac{XiL} virtualisation levels, in part by utilising LExCI's \ac{TL} capabilities.
The resulting strategy led to lower NO\textsubscript{x} and soot emissions while
maintaining the same performance as a virtual and a real \ac{ECU}.
Similarly, \cite{PICERNO20238266} applied the framework to learn a control strategy for the 
\ac{VGT} in the same setup which likewise achieved reductions in emissions 
and better performance than the reference.
In \cite{vehicles5030050}, LExCI was embedded into a cloud-based service 
in order to train an agent to control the longitudinal acceleration 
of an electric vehicle.

To the best of our knowledge, there are no comparable solutions for bringing \ac{RL} and embedded devices together. The
only close contribution is \cite{9376968} where the authors present a conceptually similar toolchain that employs a
modified version of keras-rl's \cite{plappert2016kerasrl} \ac{DDPG}
implementation to train an agent that is executed on a \ac{RCP} system. Their program is designed such that it could 
interface various algorithm implementations from different libraries and it requires the third-party tool
ControlDesk\footnote{\url{https://www.dspace.com/de/gmb/home/products/sw/experimentandvisualization/controldesk.cfm}} to 
access the embedded system. The \acp{NN} on the embedded side were hand-coded by the authors in 
MATLAB/Simulink\footnote{\url{https://www.mathworks.com/products/simulink.html}} and are limited to fully-connected feed-
forward networks. In comparison, LExCI offers more flexibility regarding the control software, design of \acp{NN}, and the
choice of \ac{RL} algorithms.

The remainder of this paper is structured as follows: First, the foundations of
\ac{RL} and two state-of-the-art \ac{RL} algorithms are expounded in 
Sec. \ref{sec:theoretical-background}. After describing LExCI and its inner workings in
Sec. \ref{sec:software}, Sec. \ref{sec:experiments} summarises the experiments that were
conducted to showcase the viability of the framework and discusses the results. Finally, Sec. \ref{sec:conclusion} recapitulates LExCI's performance,
its strengths, and how it can be extended in the future.

\section{Theoretical Background}
\label{sec:theoretical-background}

In order to understand the manner in which LExCI operates, it is crucial to cover the theory behind \ac{RL}. Along with the general concepts, this section delineates two state-of-the-art algorithms and their distinct requirements regarding the framework.

\subsection{Reinforcement Learning}
\label{subsec:reinforcement-learning}
\Ac{RL} is a \ac{ML} paradigm based on the concept of training an agent by letting it freely 
interact with its environment. The experiences that are generated in the process are 
collected and utilised to update the agent's policy such that the cumulated reward 
it receives for its behaviour is maximised. \cite{SuttonBarto}

The mathematical foundation of the environment is a time-discrete \ac{MDP} defined by the 
four-tuple $(S, A, P, R)$, that is
\begin{itemize}
  \item the set of all possible states $S$,
  \item the action space $A$,
  \item the transition probability function $P: S \times A \times S \to [0, 1]$, and
  \item the reward function $R: S \times A \times S \to \mathbb{R}$.
\end{itemize}
During an interaction, the agent observes the current state $s_{t} \in S$ and chooses an 
action $A \ni a_{t} \sim \pi_{\theta}(\cdot | s_{t})$. This causes the environment 
to transition into  the next state $s'_{t} = s_{t + 1} \in S$ with a probability of
$P(s'_{t} | s_{t}, a_{t})$ and the reward $r_{t} = R(s_{t}, a_{t}, s'_{t})$ is given. The 
flag $d$ indicates whether $s'$ is a terminal state ($d = 1$) or not ($d = 0$).
The action distribution $\pi_{\theta}: S \times A \to [0, 1]$ with configurable parameters 
$\theta$ is the agent's policy and determines its strategy. An episode or trajectory is a 
sequence $\tau = (\chi_{0}, \chi_{1}, \dots, \chi_{T})$
of experiences $\chi_{t} = (s_{t}, a_{t}, s'_{t}, r_{t}, d_{t})$. The goal of \ac{RL} is to tweak $\theta$ in order to
maximise the discounted return with a discount factor $\gamma$ (Eq. \ref{eq:discounted-return}) or the expected return (Eq. \ref{eq:expected-return}).

\begin{equation}
  \label{eq:discounted-return}
  R(\tau) = \sum_{t = 0}^{T}{\gamma^{t} r_{t}}, \quad \gamma \in (0, 1]
\end{equation}

\begin{equation}
  \label{eq:expected-return}
  J(\pi_{\theta}) = \mathbb{E}_{\tau \sim \pi_{\theta}}[R(\tau)]
\end{equation}

One prominent optimisation 
method is gradient ascent which performs iterative update steps
\begin{equation}
  \label{eq-gradient-ascent}
  \theta_{i + 1} = \theta_{i} + \eta \nabla_{\theta}{J(\pi_{\theta_{i}})}
\end{equation}
with a learning rate $\eta \in \mathbb{R}$.
The policy is typically implemented as an \ac{NN} in which case the parameter set $\theta$ 
consists of its weights and biases. \cite{SuttonBarto}

There are three key metrics to quantify how well an agent fares in a certain situation:
The \ac{VF} $V_{\pi_{\theta}}$ (Eq. \ref{eq-value-function}) estimates the return at a state 
$s \in S$ when acting on-policy (i.e. when choosing actions according to the current policy) from there on. Similarly, the action-value 
function or Q-function $Q_{\pi_{\theta}}$ (Eq. \ref{eq-action-value-function}) estimates the return when taking 
an action $a \in A$ at a state $s \in S$ on the assumption that all following actions are on-policy. The advantage function $A_{\pi_{\theta}}$ (Eq. \ref{eq-advantage-function}) is the difference of 
the two and measures how much better it is to take an action compared to what the 
policy would do. \cite{SuttonBarto}
\begin{equation}
  \label{eq-value-function}
  V_{\pi_{\theta}}(s) = \mathbb{E}_{\tau \sim \pi_{\theta}}{\left[ R(\tau) | s_{0} = s \right]}
\end{equation}
\begin{equation}
  \label{eq-action-value-function}
  Q_{\pi_{\theta}}(s, a) = \mathbb{E}_{\tau \sim \pi_{\theta}}{\left[ R(\tau) | s_{0} = s, 
  a_{0} = a \right]} = r + \gamma V_{\pi_{\theta}}(s')
\end{equation}
\begin{equation}
  \label{eq-advantage-function}
  A_{\pi_{\theta}}(s, a) = Q_{\pi_{\theta}}(s, a) - V_{\pi_{\theta}}(s)
\end{equation}
Approximations of the above are denoted as
$\hat{V}_{\pi_{\theta}}$, $\hat{Q}_{\pi_{\theta}}$, and $\hat{A}_{\pi_{\theta}}$, respectively.

An important property that distinguishes \ac{RL} algorithms is whether they insist that 
the actions in their training data be sampled using the current policy. Those that do are 
called \emph{on-policy}, the rest \emph{off-policy}. Furthermore, if the algorithm has access 
to a model of the environment or learns one for the purpose of predicting the outcome of 
actions, it is called \emph{model-based}, otherwise \emph{model-free}. It has to be noted that this model is distinct from the agent's behaviour model or any of its value function approximators. \cite{SuttonBarto, SpinningUpKindsOfRLAlgorithms}

\subsection{Algorithms}
\label{subsec:algorithms}

\subsubsection{Proximal Policy Optimization}
\label{subsubsec:ppo} 
\Ac{PPO} is a state-of-the-art model-free, on-policy \ac{RL} algorithm for discrete and continuous 
action spaces. It features a surrogate loss function whose scaled advantages are clipped to avoid excessively large update 
steps that could destabilise the training. To that end, \ac{PPO} trains a \ac{VF} approximator in addition to the 
policy. \cite{DBLP:journals/corr/SchulmanWDRK17, SpinningUpPPO}

The algorithm first garners a train batch, i.e. a defined number of experiences, using its current parameters $\theta$. When updating the agent, subsets known as mini-batches are drawn therefrom to perform multiple steps of \ac{SGD} to minimise the following loss function:
\begin{equation}
  \label{eq-ppo-surrogate-loss}
  L_{\theta}(\chi, \xi) = - \left( L_{\theta}^{\text{clip}}(\chi, \xi) + L_{\theta}^{\text{KL}}(\chi, \xi) - c_{\text{VF}}L_{\theta}^{\text{VF}}(\chi) + c_{\text{S}}S(\chi, \xi) \right)
\end{equation}
$\xi$ denotes the policy's parameter set after a \ac{SGD} step. The individual components of Eq. \ref{eq-ppo-surrogate-loss} are the clipped surrogate objective
\begin{equation}
  \label{eq-ppo-clip-loss}
  \begin{split}
    L_{\theta}^{\text{clip}}(\chi, \xi) = & \min\left( \max\left( \min\left( \frac{\pi_{\xi}(a | s)}{\pi_{\theta}(a | s)}, 1 + \epsilon \right), 1 - \epsilon \right) \hat{A}_{\pi_{\theta}}(s, a), \right. \\
    & \left. \frac{\pi_{\xi}(a | s)}{\pi_{\theta}(a | s)}\hat{A}_{\pi_{\theta}}(s, a) \right)
  \end{split}
\end{equation}
for a clip parameter $\epsilon \in \mathbb{R}$, the \ac{KL} divergence penalty
\begin{equation}
  \label{eq-ppo-kl-loss}
  L_{\theta}^{\text{KL}}(\chi, \xi) = \frac{\pi_{\xi}(a | s)}{\pi_{\theta}(a | s)}\hat{A}_{\pi_{\theta}}(s, a) - \beta \cdot \operatorname{KL}(\pi_{\theta}({} \cdot {} | s), \pi_{\xi}({} \cdot {} | s))
\end{equation}
with an adaptive coefficient $\beta \in \mathbb{R}$, the squared error of the \ac{VF} approximator
\begin{equation}
  L_{\theta}^{\text{VF}}(\chi) = \left( \hat{V}_{\pi_{\theta}}(s) - V_{\pi_{\theta}}(s) \right)^{2}
\end{equation}
and its coefficient $c_{\text{VF}} \in \mathbb{R}$, and an optional entropy bonus $S(\chi, \xi)$ 
with its coefficient $c_{\text{S}} \in \mathbb{R}$ to encourage exploration. \cite{DBLP:journals/corr/SchulmanWDRK17, SpinningUpPPO}

\subsubsection{Deep Deterministic Policy Gradient}
\label{subsubsec:ddpg}
The \ac{DDPG} algorithm is a modern model-free, off-policy \ac{RL} method which extends the idea 
of the \ac{DQN} algorithm to continuous action spaces. Since its policy is deterministic, exploration is achieved by adding random noise, e.g. from a Gaussian distribution or an \ac{OU} process, to its output. \cite{lillicrap2019continuous, SpinningUpDDPG}

\Ac{DDPG} trains a \ac{NN} with parameters $\theta_{Q}$ as an approximation 
$\hat{Q}_{\theta_{Q}}$ of the Q-function when acting greedily and another NN with parameters 
$\theta_{\mu}$ for the deterministic policy $\mu_{\theta_{\mu}}$ that seeks to maximise $\hat{Q}_{\theta_{Q}}$. To stabilise training, target 
networks $\hat{Q}_{\theta'_{Q}}$ and $\mu_{\theta'_{\mu}}$ with parameters $\theta'_{Q}$ and 
$\theta'_{\mu}$ are employed.
The Q-network is trained by minimising
\begin{equation}
  \label{eq-ddpg-q-loss}
  L_{\theta_{Q}, \theta'_{Q}, \theta'_{\mu}}(\chi) = \left( \hat{Q}_{\theta_{Q}}(s, a) - \left( r + \gamma(1 - d)\hat{Q}_{\theta'_{Q}}(s', \mu_{\theta'_{\mu}}(s')) \right) \right)^{2}
\end{equation}
and the policy is updated by performing gradient ascent using 
$\nabla_{\theta_{\mu}}\hat{Q}_{\theta_{Q}}(s, \mu_{\theta_{\mu}}(s))$. \cite{lillicrap2019continuous, SpinningUpDDPG}

The target networks are updated via polyak averaging, i.e.
\begin{equation}
  \label{eq-ddpg-polyak-averaging-q-target}
  \theta'_{Q, i + 1} = \rho \theta_{Q, i} + (1 - \rho) \theta'_{Q, i}
\end{equation}
\begin{equation}
  \label{eq-ddpg-polyak-averaging-mu-target}
  \theta'_{\mu, i + 1} = \rho \theta_{\mu, i} + (1 - \rho) \theta'_{\mu, i}
\end{equation}
for $\rho \ll 1$. Also, batches are sampled from a replay memory buffer which can be supplemented with off-policy experiences. \cite{lillicrap2019continuous, SpinningUpDDPG}

\section{Software}
\label{sec:software}

This section describes LExCI's components, how it operates, and the steps one has to take in order to set it up for a new \ac{RL} problem. Furthermore, the \emph{RL Block}, a plug-and-play Simulink model that encapsulates all necessary parts to execute an agent's policy model and to store experiences, is presented.

\subsection{Architecture and General Workflow}
\label{subsec:software:architecture-and-general-workflow}
\begin{figure}
  \centering
  \includegraphics[scale=0.275]{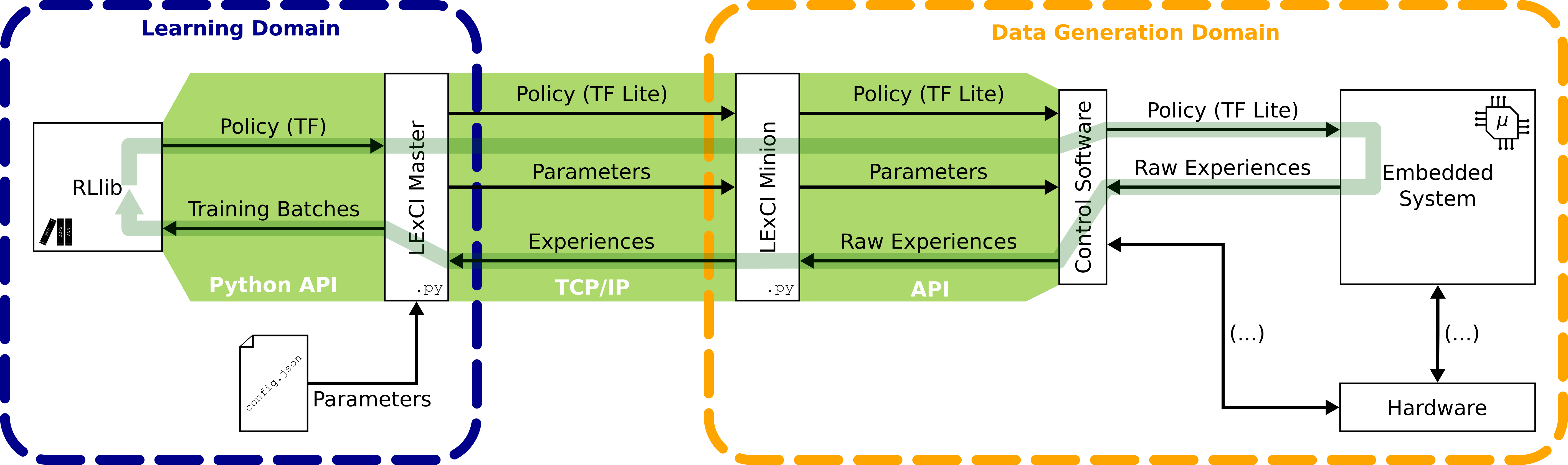}
  \caption{Software architecture of the LExCI framework with the eponymous cycle as a light green arrow. There are multiple independent instances of the data generation domain when the process is parallelised.}
  \label{fig:lexci-architecture}
\end{figure}

\begin{figure}[b]
  \centering
  \includegraphics[scale=0.15]{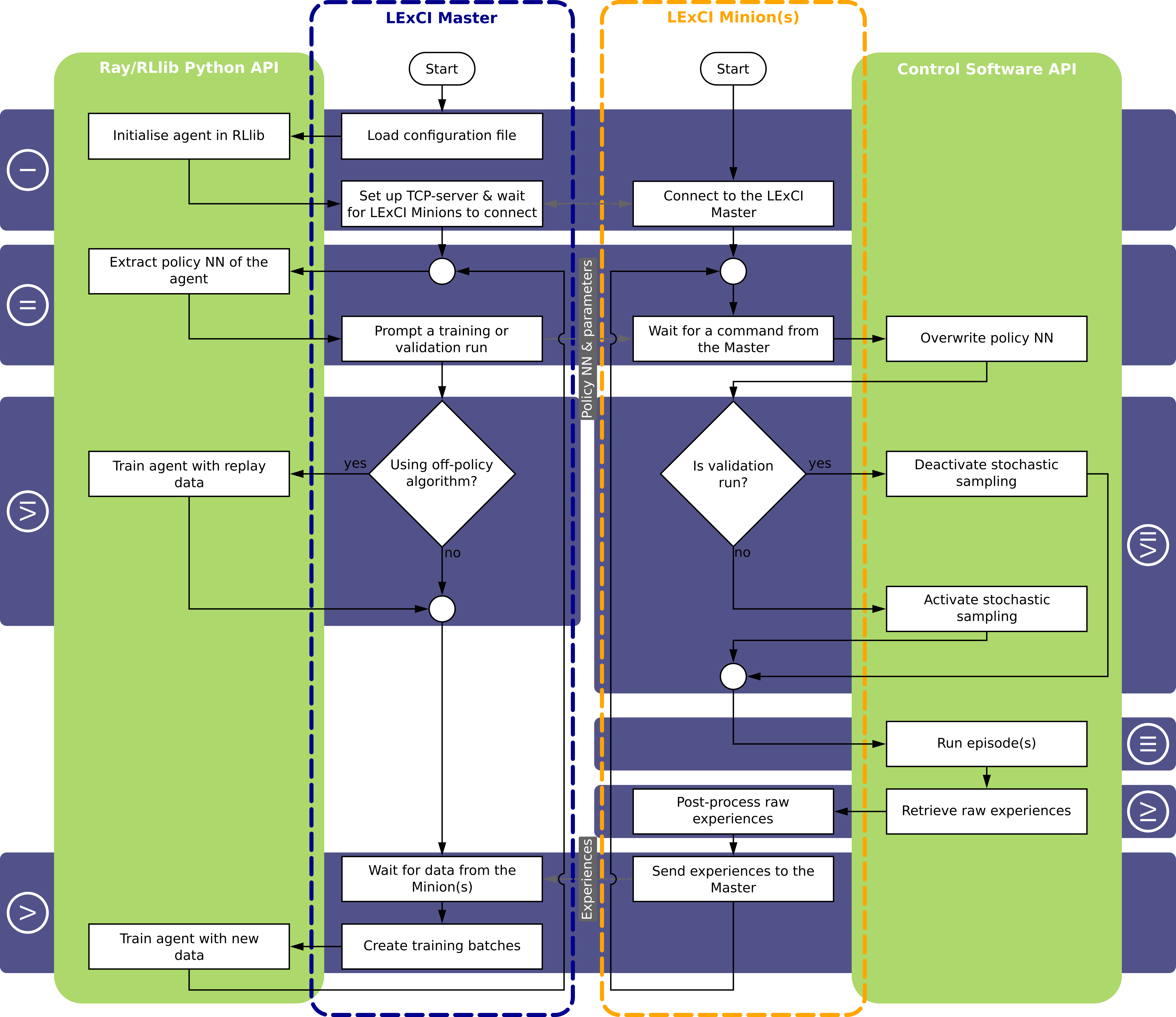}
  \caption{Simplified flowchart of the LExCI framework. The grey, dashed arrows indicate communication/data exchange between the Minion and the Master. The blue areas tagged with Roman numerals serve as references for the textual description of the figure.}
  \label{fig:lexci-flowchart}
\end{figure}

LExCI is logically divided into two domains as illustrated in Fig. \ref{fig:lexci-architecture}: 
The first is the learning side of the framework with the LExCI Master at its head.
Its counterpart is the data generation side where the LExCI Minion is located.
To understand their roles and how they work together, it is best to have a look at the framework's modus operandi. As an aid, Fig. \ref{fig:lexci-flowchart} complements Fig. \ref{fig:lexci-architecture} with the chronological order of the steps. The sections highlighted there shall be used as a guide.

\textbf{Section I} The LExCI Master makes use of a slightly modified version of Ray/RLlib 1.13.0\footnote{The modified version allows \ac{DDPG} agents to choose between on-policy/off-policy training and is part of the LExCI repository.} via the library's Python \ac{API}.
At program startup, it loads a JSON-formatted configuration file containing the parameters of the training.
These include the characteristics of the problem (the dimensions of the observation and action space, whether actions are continuous or discrete, etc.), general settings (networking details, where to store logs and results, and the like) as well as the algorithm's hyperparameters (the architecture of the agent's \ac{NN}(s), the \ac{LR}/schedule, or batch sizes to name a few).
The Master initialises the agent based on the settings above before proceeding to its main loop for training.
In addition to being the gateway to the \ac{RL} library, the Master acts as a server and listens for incoming TCP/IP connections from LExCI Minions.
Established connections are constantly monitored for their status and closed if the opposite side stops sending heartbeats (e.g. after a program crash) or takes too long to finish its task.
Thus, the system is able to cope with unforeseen events.

\textbf{Sections II \& III} Training is carried out by completing so called cycles.
At the beginning of a cycle, the LExCI Master retrieves the agent's current policy from RLlib and converts it from its original \ac{TF} format to \ac{TF Lite}.
This model, along with all relevant training parameters (e.g. the number of experiences to generate), is broadcast to the connected LExCI Minions using a custom JSON-based protocol.
Upon receipt, each Minion utilises the \ac{API} of its control software to overwrite the policy on the embedded device which is then prompted to generate experiences.
Additional pieces of hardware can be part of the data generation domain and interact with the embedded device. Besides the closed-loop control system, those include physical actuators or sensors.

\textbf{Section IV} Once enough data has been collected, the Minion uses the control software again to get the raw experiences and post-processes them.
For instance, a domain expert could define auxiliary penalties that are added to the reward in situations where the agent's actions were clearly nonsensical.

\textbf{Section V} The experiences are sent to the LExCI Master and arranged into training batches, i.e. the data format RLlib expects for training.
During that process, experiences are supplemented with additional information if the algorithm calls for it.
For example, \ac{PPO} requires the predicted value of the \ac{VF} approximator (see Eq. \ref{eq-value-function}), the action distribution, and the probability of the action on top
of the standard quantities.
After the training batch has been assembled, it is given to RLlib for training the agent and the cycle starts anew.

\textbf{Section VI} When learning with off-policy algorithms, the LExCI Master does not remain idle while the Minions are doing their part.
Instead, the Master continues training with experiences drawn from its replay memory buffer.
The size of the buffer, the number of replay training steps per cycle, and the extent to which the buffer must be filled before replay training starts 
are set in the configuration file.

\textbf{Section VII} Apart from training runs, LExCI can be configured to conduct validation episodes with a defined frequency.
They differ in that actions are always set to the mean of the action distribution and are hence deterministic during validations rather than being sampled stochastically.
Thus, the results are more comparable and lend themselves better to assessing the agent's performance.
Further, validations are conducted by a single Minion.

The master-minion architecture has the added benefit that it enables easy parallelisation.
In light of the fact that embedded devices usually operate in real-time, this feature can speed up the data generation process dramatically.
When there are multiple LExCI Minions available, the Master splits the workload between them so each only has to generate a fraction of the required number of experiences.

\subsection{Setup}
\label{subsec:software-setup}
When employing the framework for a new use-case, the Master and the Minion must first be set up.
LExCI is shipped with what is called a \emph{universal Master} for each \ac{RL} 
algorithm.
Those are ready-to-use Python programs that function as described in Sec. \ref{subsec:software:architecture-and-general-workflow}, so one merely has to select the right algorithm and adjust the parameters in the configuration file.
Alternatively, users can write their own custom Master programs which create an instance of the \lstinline[basicstyle=\ttfamily]{Master} class and call its main loop.
The Minion is always tailor-made for the problem by writing a program that instantiates the \lstinline[basicstyle=\ttfamily]{Minion} class and invokes its main loop.
There, the logic for preparing the embedded system, overwriting the agent's model, running episodes, post-processing experiences, etc. is programmed.
The class expects callback functions for generating training and validation data.
To this effect, LExCI offers helper classes that facilitate interacting with the embedded system via a control software.
At the time of writing, there are helpers for ControlDesk, MATLAB/Simulink, and ECU-TEST\footnote{\url{https://www.tracetronic.com/products/ecu-test/}}.

Another significant facet of the setup process involves the software that shall be 
running on the embedded system itself. After all, it is responsible for 
executing the policy \ac{NN} of the agent. Users are free to 
implement the inference of actions in whatever way they deem fit. Having said that, it is of paramount 
importance that they distinguish between what are called \emph{normalised} and \emph{denormalised} spaces. 
Normalised observations and actions are the raw quantities passed to and received from \acp{NN}. 
Denormalised quantities, on the other hand, are the ones that the environment provides or expects.
It is standard practice to, for example, min-max normalise observations (from the environment) to the range 
$[-1, +1]$ (which would then be the agent's normalised observation space) to stabilise and expedite 
training \cite{SuttonBarto}. By the same token, the normalised actions 
of the agent  must be mapped to the allowed 
(denormalised) range in the environment, e.g. via a hyperbolic tangent 
and scaling or simply by clipping. The data that the Minion retrieves 
from the embedded system must always be normalised.
To aid users, LExCI comes with software modules that can be used to 
execute the agent (\lstinline[basicstyle=\ttfamily]{neural_network_module}) and to transform quantities between the spaces.

\begin{figure}[h!]
  \centering
  \includegraphics[scale=0.75]{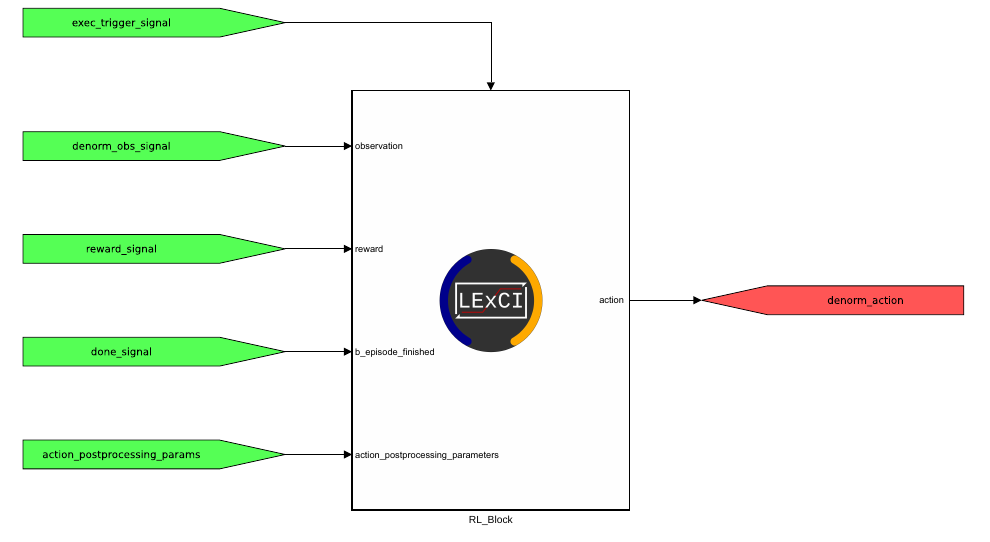}
  \caption{LExCI's RL Block in Simulink. The ports \lstinline[basicstyle=\ttfamily]{observation} and \lstinline[basicstyle=\ttfamily]{action} are in the denormalised space of the environment.}
  \label{fig:lexci-rl-block}
\end{figure}

Considering how widely used MATLAB/Simulink are in the engineering domain, especially in 
control prototyping, LExCI's \emph{RL Block} (Fig. \ref{fig:lexci-rl-block}) plays a prominent 
role in that regard. It is a ready-to-use Simulink subsystem that houses the \ac{RL}-based 
controller such that employing it becomes as simple as copying it into the plant model, connecting its ports, and setting some basic parameters.
Inside, the RL Block min-max normalises observations, feeds them to the policy 
\ac{NN} of the agent, samples an action from the inferred action distribution, and 
denormalises the same before returning it. Its centrepieces are the S-Function containing the 
C++-code to execute the agent using \ac{TFLM} and the internal experience buffer which can be 
accessed via the control software. The RL Block is externally triggered so that the agent can be executed at a different (i.e. slower) sample rate than the surrounding model.

\section{Experiments}
\label{sec:experiments}
For this paper, LExCI was applied to the inverted pendulum 
swing-up problem which is a standard benchmark for continuous control.
To highlight its versatility, multiple trainings were performed with the 
framework, each with a different \ac{RL} algorithm and target system.

\subsection{Pendulum Environment and Setup}
\label{subsec:pendulum-environment-and-setup}
\begin{figure}[h]
  \centering
  \includegraphics[scale=2.0]{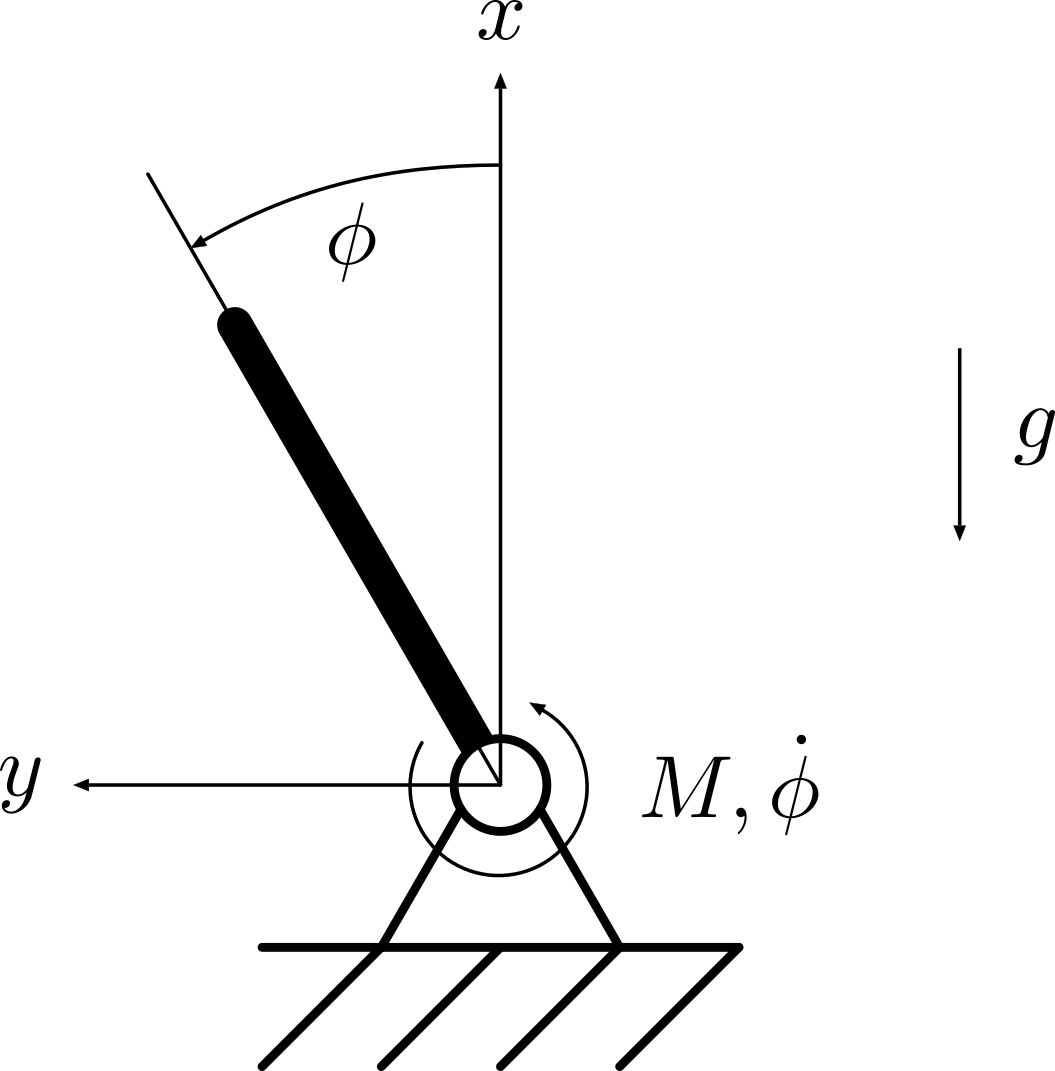}
  \caption{The pendulum swing-up problem according to \cite{gymnasium-pendulum}.}
  \label{fig:pendulum-swingup-problem}
\end{figure}

In the inverted pendulum 
swing-up environment, a rod of length $l = \SI{1}{\meter}$ and 
mass\footnote{We chose $g = \SI{9.81}{\meter\per\second\squared}$ instead of the default value of $10$ in \cite{openai-pendulum-implementation}.}
$m = \SI{1}{\kg}$ is mounted to a wall on one end with a single rotational 
degree of freedom (cf. Fig. \ref{fig:pendulum-swingup-problem}). The 
objective is to apply a torque $M$ at the pivot point in every time step 
such that it stands upright, i.e. the angle $\phi \in (-\pi, +\pi]$ 
between the rod and the vertical axis as well as its angular velocity 
$\dot{\phi}$ become $0$. The time step length is $\Delta t = \SI{0.05}{\second}$. Using 
$x = l \cdot \cos(\phi)$ and $y = l\cdot \sin(\phi)$, Tab. 
\ref{tab:pendulum-swingup-problem-observation-action-space} summarises the 
environment's observation and action space while Eq. \ref{eq:pendulum-swingup-problem-reward-function}
describes its reward function. Episodes are $200$ time steps long and start 
at a random position $\phi_{0} \in (-\pi, +\pi]$ and with a random 
angular velocity $\dot{\phi}_{0} \in [\SI{-1}{\radian \per \second}, \SI[retain-explicit-plus]{+1}{\radian \per \second}]$. \cite{gymnasium-pendulum, openai-pendulum-implementation, bi2021deep}

\begin{equation}
  \label{eq:pendulum-swingup-problem-reward-function}
  R(\phi, \dot{\phi}, M) = -\phi^{2} - 0.1 \cdot \dot{\phi}^2 - 0.001 \cdot M^{2}
\end{equation}

\begin{table}[h]
  \centering
  \begin{tabular}{| c | c | c | c | c |}
    \hline
    Number & Observation & Minimum & Maximum & Unit \\
    \hline
    1 & $x$ & $-1$ & $+1$ & $\si{\meter}$ \\
    2 & $y$ & $-1$ & $+1$ & $\si{\meter}$ \\
    3 & $\dot{\phi}$ & $-8$ & $+8$ & $\si{\radian \per \second}$ \\
    \hline
    \hline
    Number & Action & Minimum & Maximum & Unit \\
    \hline
    1 & $M$ & $-2$ & $+2$ & $\si{\newton \meter}$ \\
    \hline
  \end{tabular}
  \caption{The observation and action space of the pendulum swing-up problem.}
  \label{tab:pendulum-swingup-problem-observation-action-space}
\end{table}

\pagebreak
The pendulum problem was tackled three times with LExCI:
\begin{description}
    \item[\textbf{Python}] First, purely in Python using the gym
                  implementation of the environment \cite{openai-pendulum-implementation}
                  and LExCI's \lstinline[basicstyle=\ttfamily]{neural_network_module}s
                  (cf. Sec. \ref{subsec:software-setup}) to execute the agent's policy.
    \item[\textbf{Simulink}] Second, with the pendulum environment running in Simulink using the RL Block
                    (see Sec. \ref{subsec:software-setup}) and a custom
                    model that is identical in behaviour to gym's implementation.
    \item[\textbf{MABX III}] Third, with the environment running on a dSPACE
                    \ac{MABX} III\footnote{\url{https://www.dspace.com/en/pub/home/products/hw/micautob/microautobox3.cfm}},
                    a \ac{RCP} system commonly used for embedded control by the automotive industry. This run, too, used a
                    custom model of the pendulum environment and the RL Block
                    (cf. Sec. \ref{subsec:software-setup}).
\end{description}

With each target system, one agent 
was trained with \ac{PPO} and one with \ac{DDPG} for a total of six 
training runs. The choice of algorithms was motivated by their wide-spread use in engineering and the fact that one is on-policy while the other is not. Observations were min-max normalised and the real-valued actions were mapped via a scaled hyperbolic tangent to the boundaries of the environment (see Sec. \ref{subsec:software-setup}).
The hyperparameters were chosen based on RLlib's pre-tuned configurations for the respective algorithms and extended by LExCI's custom ones. App. \ref{sec:appendix-hyperparameters} lists the most important parameters.

Validations were performed every five cycles so that the agent's performance was tested frequently enough without
creating too much overhead. For that purpose, the pendulum environment was initialised with $\phi_{0} = \pi$ and 
$\dot{\phi}_{0} = \SI{0}{\radian \per \second}$, i.e. with the rod hanging still at the six o'clock position.

\subsection{Results}
\label{subsec:results}
Given the definition of the pendulum environment and the hyperparameters that were 
chosen, three episodes were generated in every cycle.
Fig. \ref{fig:ppo-average-training-returns-smoothed} and
Fig. \ref{fig:ddpg-average-training-returns-smoothed} plot their smoothed average 
returns over the cycle number while the unfiltered quantities can be found in Fig. 
\ref{fig:ppo-average-training-returns} and Fig. \ref{fig:ddpg-average-training-returns} 
of App. \ref{sec:appendix-additional-training-data}.
The plots show some noteworthy characteristics of the trainings: First, every 
combination of \ac{RL} algorithm and target system converged towards the optimum 
where the agent exhibits good performance. To substantiate this claim, 
Fig. \ref{fig:ddpg-mabx-validation-45} shows the best validation run of the \ac{DDPG}-training on the \ac{MABX} III where the agent swings the pendulum to the 12 o'clock position
($x = \SI{1}{\meter}$ and $y = \SI{0}{\meter}$) within the first 50 time steps (i.e. in just $\SI{2.5}{\second}$) and holds 
it there for the remainder of the episode ($\dot{\phi} = \SI{0}{\radian \per \second}$). The same is true for the best \ac{PPO} validation on that platform (cf. Fig. \ref{fig:ppo-mabx-validation-750}).
Other combinations performed analogously once the training had converged (see Fig. \ref{fig:ppo-python-validation-745}, Fig. \ref{fig:ddpg-python-validation-45}, Fig. \ref{fig:ppo-simulink-validation-745}, and Fig. \ref{fig:ddpg-simulink-validation-45} in App. \ref{sec:appendix-additional-training-data}). Please note that the variations in maximum return are merely a result of the stochastic nature of exploration paired with the random initialisation of the environment at the beginning of every episode. They do not mean that one target system performs better than the others.
Second, all target systems display a similar course of training progression 
for each algorithm which proves that i) our Simulink and \ac{MABX} III models 
of the pendulum environment are equal to gym's implementation in terms of 
behaviour and ii) that LExCI is able to train well on various platforms. 
Furthermore, Fig. \ref{fig:ppo-average-training-returns-smoothed} and Fig. \ref{fig:ddpg-average-training-returns-smoothed}
are in accord with the results of \cite{kumar2021controlling} though the author used a different set of hyperparameters.
Third, all agents remain stable after convergence. The oscillations in the average cycle returns are mainly caused by the random initialisation of the pendulum which sometimes starts in more and sometimes in less advantageous states.

\begin{figure}[h!]
  \begin{minipage}[b]{0.475\textwidth}
  \centering
  \begin{tikzpicture}[scale=0.675]
    \definecolor{pythoncolor}{RGB}{93,130,43};
    \definecolor{simulinkcolor}{RGB}{192,76,11};
    \definecolor{mabxcolor}{RGB}{0,68,148};
  
    \begin{axis}[
        grid=major,
        xlabel=Cycle / -,
        y label style={at={(axis description cs:-0.175,0.5)}},
        ylabel=Return / -,
        legend pos=north west
    ]
      \addplot[color=pythoncolor] table [x="Cycle", y="Python", col sep=semicolon] {PPO_Average_Training_Returns_Smoothed.csv};
      \addlegendentry{Python};
      \addplot[color=simulinkcolor] table [x="Cycle", y="Simulink", col sep=semicolon] {PPO_Average_Training_Returns_Smoothed.csv};
      \addlegendentry{Simulink};
      \addplot[color=mabxcolor] table [x="Cycle", y="MABX III", col sep=semicolon] {PPO_Average_Training_Returns_Smoothed.csv};
      \addlegendentry{MABX III};
    \end{axis}
  \end{tikzpicture}
  \caption{Average LExCI PPO training returns with three episodes per cycle. The data has been smoothed with a moving average filter of size 11.}
  \label{fig:ppo-average-training-returns-smoothed}
  \end{minipage}
  \hspace{0.05\textwidth}
  \begin{minipage}[b]{0.475\textwidth}
  \centering
  \begin{tikzpicture}[scale=0.675]
    \definecolor{pythoncolor}{RGB}{93,130,43};
    \definecolor{simulinkcolor}{RGB}{192,76,11};
    \definecolor{mabxcolor}{RGB}{0,68,148};
  
    \begin{axis}[
        grid=major,
        xlabel=Cycle / -,
        y label style={at={(axis description cs:-0.175,0.5)}},
        ylabel=Return / -,
        legend pos=north west
    ]
      \addplot[color=pythoncolor] table [x="Cycle", y="Python", col sep=semicolon] {DDPG_Average_Training_Returns_Smoothed.csv};
      \addlegendentry{Python};
      \addplot[color=simulinkcolor] table [x="Cycle", y="Simulink", col sep=semicolon] {DDPG_Average_Training_Returns_Smoothed.csv};
      \addlegendentry{Simulink};
      \addplot[color=mabxcolor] table [x="Cycle", y="MABX III", col sep=semicolon] {DDPG_Average_Training_Returns_Smoothed.csv};
      \addlegendentry{MABX III};
    \end{axis}
  \end{tikzpicture}
  \caption{Average LExCI DDPG training returns with three episodes per cycle. The data has been smoothed with a moving average filter of size 11.}
  \label{fig:ddpg-average-training-returns-smoothed}
  \end{minipage}
\end{figure}

\begin{figure}[h!]
  \begin{minipage}[b]{0.475\textwidth}
  \centering
  \begin{tikzpicture}[scale=0.675]
    \definecolor{xcolor}{RGB}{0,70,135};
    \definecolor{ycolor}{RGB}{255,70,15};
    \definecolor{phidotcolor}{RGB}{0,90,0};
  
    \begin{axis}[
        grid=major,
        xlabel={Time Step / - ($\Delta t = \SI{0.05}{\second}$)},
        ymin=-9, ymax=9,
        ytick distance=2,
        legend pos=north east
    ]
      \addplot[color=xcolor] table [x="Time Step", y="x", col sep=semicolon] {PPO_MABX_Validation_Cycle_750.csv};
      \addlegendentry{$x$ / $\SI{}{\meter}$}
      \addplot[color=ycolor] table [x="Time Step", y="y", col sep=semicolon] {PPO_MABX_Validation_Cycle_750.csv};
      \addlegendentry{$y$ / $\SI{}{\meter}$}
      \addplot[color=phidotcolor] table [x="Time Step", y="Angular Velocity", col sep=semicolon] {PPO_MABX_Validation_Cycle_750.csv};
      \addlegendentry{$\dot{\phi}$ [$\SI{}{\radian\per\second}$]}
    \end{axis}
  \end{tikzpicture}
  \caption{Best validation at cycle 750 of the LExCI PPO training with the \ac{MABX} III. The return of the episode was $-378.22$.}
  \label{fig:ppo-mabx-validation-750}
  \end{minipage}
  \hspace{0.05\textwidth}
  \begin{minipage}[b]{0.475\textwidth}
  \centering
  \begin{tikzpicture}[scale=0.675]
    \definecolor{xcolor}{RGB}{0,70,135};
    \definecolor{ycolor}{RGB}{255,70,15};
    \definecolor{phidotcolor}{RGB}{0,90,0};
  
    \begin{axis}[
        grid=major,
        xlabel={Time Step / - ($\Delta t = \SI{0.05}{\second}$)},
        ymin=-9, ymax=9,
        ytick distance=2,
        legend pos=north east
    ]
      \addplot[color=xcolor] table [x="Time Step", y="x", col sep=semicolon] {DDPG_MABX_Validation_Cycle_45.csv};
      \addlegendentry{$x$ / $\SI{}{\meter}$}
      \addplot[color=ycolor] table [x="Time Step", y="y", col sep=semicolon] {DDPG_MABX_Validation_Cycle_45.csv};
      \addlegendentry{$y$ / $\SI{}{\meter}$}
      \addplot[color=phidotcolor] table [x="Time Step", y="Angular Velocity", col sep=semicolon] {DDPG_MABX_Validation_Cycle_45.csv};
      \addlegendentry{$\dot{\phi}$ / $\SI{}{\radian\per\second}$}
    \end{axis}
  \end{tikzpicture}
  \caption{Best validation at cycle 45 of the LExCI DDPG training with the \ac{MABX} III. The return of the episode was $-367.32$.}
  \label{fig:ddpg-mabx-validation-45}
  \end{minipage}
\end{figure}

To further validate the results, the pendulum environment 
was also trained without LExCI, i.e. using Ray/RLlib only. These trainings shall be referred to as \emph{native}. For the sake of comparability, 
the environment was configured such that observations are min-max normalised and actions 
are mapped with a scaled hyperbolic tangent.
When analysing the results in Fig. \ref{fig:ppo-native-average-training-returns}
and Fig. \ref{fig:ddpg-native-average-training-returns} and comparing 
them to the ones above, one has to consider two things: 1) Ray/RLlib's iterations do not directly
correspond to LExCI's cycles. Because of that, the hyperparameters from App. \ref{sec:appendix-hyperparameters} 
had to be slightly varied so as to best replicate LExCI's behaviour. This mainly affected the \ac{DDPG} settings that govern how many samples are generated and how often replay data is used for training. 2) Native Ray/RLlib utilises an \ac{OU} process for 
exploration and not Gaussian noise for \ac{DDPG}. With that in mind, the average training returns 
display the same general progress and --- more important --- have the same 
minima and maxima. This proves that LExCI interfaces RLlib correctly
and that the framework is able to train agents to the same level of 
quality as the original library setup.

\begin{figure}[h!]
  \begin{minipage}[b]{0.475\textwidth}
  \centering

  \begin{tikzpicture}[scale=0.675]
    \definecolor{ppocolor}{RGB}{0,70,135};
  
    \begin{axis}[
        grid=major,
        xlabel=Iteration / -,
        y label style={at={(axis description cs:-0.175,0.5)}},
        ylabel=Return / -,
    ]
      \addplot[color=ppocolor] table [x="Iteration", y="Native", col sep=semicolon] {PPO_Native_Average_Training_Returns.csv};
    \end{axis}
  \end{tikzpicture}
  \caption{Smoothed average PPO training returns in native Ray/RLlib.}
  \label{fig:ppo-native-average-training-returns}
  \end{minipage}
  \hspace{0.05\textwidth}
  \begin{minipage}[b]{0.475\textwidth}
  \centering
  \begin{tikzpicture}[scale=0.675]
    \definecolor{ddpgcolor}{RGB}{255,70,15};
  
    \begin{axis}[
        grid=major,
        xlabel=Iteration / -,
        y label style={at={(axis description cs:-0.175,0.5)}},
        ylabel=Return / -,
    ]
      \addplot[color=ddpgcolor] table [x="Iteration", y="Native", col sep=semicolon] {DDPG_Native_Average_Training_Returns.csv};
    \end{axis}
  \end{tikzpicture}
  \caption{Smoothed average DDPG training returns in native Ray/RLlib.}
  \label{fig:ddpg-native-average-training-returns}
  \end{minipage}
\end{figure}

\section{Conclusion}
\label{sec:conclusion}
This paper explained the importance of \ac{RL} for developing 
today's and tomorrow's control functions and highlighted the 
difficulties engineers face during training and deployment of \ac{RL} 
agents with/on embedded devices. The LExCI framework was presented as 
an open-source solution and its 
performance has been demonstrated across various target systems, including 
a state-of-the-art \ac{RCP} system, for a classic control task. Not 
only did LExCI succeed in integrating those platforms into the process, the 
results were also on a par with what the underlying \ac{RL} library 
can produce natively on a conventional \ac{PC}.

The framework enables users to apply \ac{RL} to real-world engineering 
problems on professional hardware as has been shown in prior works.
Considering that one had to resort to specialised solutions to
do so in the past, LExCI facilitates the process many times over because
of its generic interface to embedded devices. Moreover, the fact that it relies
on free, established libraries means that end-users are not forced to content
themselves with proprietary implementations. Instead, they can leverage the
full expertise of the open-source communities behind said libraries and thus
obtain better results.

In the future, LExCI will be updated to the latest RLlib release as 
the latter has since undergone a major version change. Additionally, 
support for more algorithms will be implemented as well as features that 
aid in exercising advanced techniques. For instance, the framework shall have 
a more extensive repertoire of \ac{TL} functionalities.

\backmatter

\bmhead{Acknowledgments}

We would like to thank TraceTronic GmbH and our student assistant Abdus Hashmy 
for their valuable support.

\section*{Declarations}
\subsection*{Funding}
This work and the scientific research behind it have been funded by the \emph{Hy-Nets4all} project 
(grant no. EFRE-0801698) of the European Regional Development Fund (ERDF), the
Federal Ministry for Economic Affairs and Climate Action (BMWK) through the
German Federation of Industrial Research Associations (AiF, IGF no. 21407 N) and assigned by the
Research Association FVV as project \emph{Heuristic Search and Deep Learning}, and the \emph{VISION} project (grant no. KK5371001ZG1) of the BMWK on the basis
of a decision by the German Bundestag.
Work was performed at the Center for Mobile Propulsion (CMP)
funded by the German Research Foundation (DFG) and the German Science and Humanities Council (WR).

\subsection*{Competing Interests}
The following could be considered a potential financial interest:
Lucas Koch, Mario Picerno, Kevin Badalian, Sung-Yong Lee, and Jakob Andert
have a patent pending for
\emph{Automatsierte Funktonskalibrierung}/\emph{Automated Feature Calibration}
(patent no. DE102022104648A1/EP4235319A1).

\subsection*{Ethics Approval}
Not applicable.

\subsection*{Consent to Participate}
Not applicable.

\subsection*{Consent for Publication}
Not applicable.

\subsection*{Availability of Data and Materials}
See \emph{Code Availability}.

\subsection*{Code Availability}
\label{subsec:code-availability}
LExCI's source code is available in its official GitHub repository:
\begin{center}
  \url{https://github.com/mechatronics-RWTH/lexci-2}
\end{center}
The version used for this paper (including the experiment code, models, and data)
can be found in the \emph{lexci\_paper} branch.

\subsection*{Authors' Contributions}
\textbf{Conceptualisation:} Kevin Badalian, Lucas Koch, Marius Wegener;
\textbf{Methodology:} Kevin Badalian, Lucas Koch, Tobias Brinkmann, Mario Picerno;
\textbf{Software:} Kevin Badalian, Lucas Koch, Tobias Brinkmann;
\textbf{Validation:} Kevin Badalian, Lucas Koch, Mario Picerno, Tobias Brinkmann;
\textbf{Formal analysis:} Kevin Badalian, Lucas Koch, Mario Picerno;
\textbf{Investigation:} Kevin Badalian, Lucas Koch;
\textbf{Resources:} Jakob Andert;
\textbf{Data curation:} Kevin Badalian, Lucas Koch, Mario Picerno;
\textbf{Writing -- original draft:} Kevin Badalian, Lucas Koch;
\textbf{Writing -- review \& editing:} Lucas Koch, Tobias Brinkmann, Mario Picerno, Marius Wegener, Sung-Yong Lee, Jakob Andert;
\textbf{Visualisation:} Kevin Badalian;
\textbf{Supervision:} Jakob Andert;
\textbf{Project administration:} Jakob Andert;
\textbf{Funding acquisition:} Jakob Andert

\bigskip

\begin{appendices}
\counterwithin{figure}{section}
\counterwithin{table}{section}
\section{Hyperparameters}
\label{sec:appendix-hyperparameters}
The hyperparameters used for training the agents presented in this paper 
are based on RLlib's pre-tuned configurations for
\ac{PPO}\footnote{\url{https://github.com/ray-project/ray/blob/ray-1.13.0/rllib/tuned_examples/ppo/pendulum-ppo.yaml}}
and \ac{DDPG}\footnote{\url{https://github.com/ray-project/ray/blob/ray-1.13.0/rllib/tuned_examples/ddpg/pendulum-ddpg.yaml}}
in the pendulum environment. Parameters not specified in Tab. \ref{tab:ppo-hyperparameters} and Tab. \ref{tab:ddpg-hyperparameters} were set to their default values.

\begin{table}[h]
  \centering
  \begin{tabular}{| l | c |}
    \hline
    Policy \ac{NN} & $\mathbf{3 \times 64 \times 64 \times 2}$, $\tanh$-activated \\ \hline
    \ac{VF} \ac{NN} & $\mathbf{3 \times 64 \times 64 \times 1}$, $\tanh$-activated \\ \hline
    Train batch size & $512$ \\ \hline
    \ac{SGD} mini-batch size & $64$ \\ \hline
    \ac{SGD} iterations per batch & $6$ \\ \hline
    $\gamma$ & $0.95$ \\ \hline
    $\lambda$ & $0.1$ \\ \hline
    $\epsilon$ & $0.3$ \\ \hline
    \ac{VF} clip parameter & \textbf{10000} \\ \hline
    \ac{LR} & 0.0003 \\ \hline
    \ac{KL} target & 0.01 \\ \hline
  \end{tabular}
  \caption{PPO-hyperparameters used for training the agents in Sec. \ref{sec:experiments}. All \acp{NN} were fully-connected and feed-forward. Values that differ from RLlib's pendulum hyperparameters are printed in bold.}
  \label{tab:ppo-hyperparameters}
\end{table}

\begin{table}[h]
  \centering
  \begin{tabular}{| l | c |}
    \hline
    Policy \ac{NN} & $3 \times 64 \times 64 \times 1$, ReLU-activated \\ \hline
    Q-function \ac{NN} & $3 \times 64 \times 64 \times 1$, ReLU-activated \\ \hline
    Replay buffer size & $10000$ \\ \hline
    Experiences per cycle$^{\ast}$ & $600$ \\ \hline
    Experiences before replay training$^{\ast}$ & $2400$ \\ \hline
    Percentage of buffer used for replay training$^{\ast}$ & $0.25$ \\ \hline
    Train batch size & $64$ \\ \hline
    $\gamma$ & $0.99$ \\ \hline
    LR (policy) & $0.001$ \\ \hline
    LR (Q-function) & $0.001$ \\ \hline
    Huber threshold & $1$ \\ \hline
    $\rho$ & $0.001$ \\ \hline
  \end{tabular}
  \caption{DDPG-hyperparameters used for training the agents in Sec. \ref{sec:experiments}. All \acp{NN} were fully-connected and feed-forward. LExCI-specific parameters are marked with an asterisk.}
  \label{tab:ddpg-hyperparameters}
\end{table}

\pagebreak
\section{Additional Training Data}
\label{sec:appendix-additional-training-data}

\begin{figure}[h!]
  \begin{minipage}[b]{0.475\textwidth}
  \centering
  \begin{tikzpicture}[scale=0.675]
    \definecolor{pythoncolor}{RGB}{93,130,43};
    \definecolor{simulinkcolor}{RGB}{192,76,11};
    \definecolor{mabxcolor}{RGB}{0,68,148};
  
    \begin{axis}[
        grid=major,
        xlabel=Cycle / -,
        y label style={at={(axis description cs:-0.175,0.5)}},
        ylabel=Return / -,
        legend pos=north west
    ]
      \addplot[color=pythoncolor] table [x="Cycle", y="Python", col sep=semicolon] {PPO_Average_Training_Returns.csv};
      \addlegendentry{Python};
      \addplot[color=simulinkcolor] table [x="Cycle", y="Simulink", col sep=semicolon] {PPO_Average_Training_Returns.csv};
      \addlegendentry{Simulink};
      \addplot[color=mabxcolor] table [x="Cycle", y="MABX III", col sep=semicolon] {PPO_Average_Training_Returns.csv};
      \addlegendentry{MABX III};
    \end{axis}
  \end{tikzpicture}
  \caption{Unfiltered average LExCI PPO training returns with three episodes per cycle.}
  \label{fig:ppo-average-training-returns}
  \end{minipage}
  \hspace{0.05\textwidth}
  \begin{minipage}[b]{0.475\textwidth}
  \centering
  \begin{tikzpicture}[scale=0.675]
    \definecolor{pythoncolor}{RGB}{93,130,43};
    \definecolor{simulinkcolor}{RGB}{192,76,11};
    \definecolor{mabxcolor}{RGB}{0,68,148};
  
    \begin{axis}[
        grid=major,
        xlabel=Cycle / -,
        y label style={at={(axis description cs:-0.175,0.5)}},
        ylabel=Return / -,
        legend pos=north west
    ]
      \addplot[color=pythoncolor] table [x="Cycle", y="Python", col sep=semicolon] {DDPG_Average_Training_Returns.csv};
      \addlegendentry{Python};
      \addplot[color=simulinkcolor] table [x="Cycle", y="Simulink", col sep=semicolon] {DDPG_Average_Training_Returns.csv};
      \addlegendentry{Simulink};
      \addplot[color=mabxcolor] table [x="Cycle", y="MABX III", col sep=semicolon] {DDPG_Average_Training_Returns.csv};
      \addlegendentry{MABX III};
    \end{axis}
  \end{tikzpicture}
  \caption{Unfiltered average LExCI DDPG training returns with three episodes per cycle.}
  \label{fig:ddpg-average-training-returns}
  \end{minipage}
\end{figure}

\begin{figure}[h!]
  \begin{minipage}[b]{0.475\textwidth}
  \centering
  \begin{tikzpicture}[scale=0.675]
    \definecolor{xcolor}{RGB}{0,70,135};
    \definecolor{ycolor}{RGB}{255,70,15};
    \definecolor{phidotcolor}{RGB}{0,90,0};
  
    \begin{axis}[
        grid=major,
        xlabel={Time Step / - ($\Delta t = \SI{0.05}{\second}$)},
        ymin=-9, ymax=9,
        ytick distance=2,
        legend pos=north east
    ]
      \addplot[color=xcolor] table [x="Time Step", y="x", col sep=semicolon] {PPO_Python_Validation_Cycle_745.csv};
      \addlegendentry{$x$ / $\SI{}{\meter}$}
      \addplot[color=ycolor] table [x="Time Step", y="y", col sep=semicolon] {PPO_Python_Validation_Cycle_745.csv};
      \addlegendentry{$y$ / $\SI{}{\meter}$}
      \addplot[color=phidotcolor] table [x="Time Step", y="Angular Velocity", col sep=semicolon] {PPO_Python_Validation_Cycle_745.csv};
      \addlegendentry{$\dot{\phi}$ / $\SI{}{\radian\per\second}$}
    \end{axis}
  \end{tikzpicture}
  \caption{Best validation at cycle 745 of the LExCI PPO training with Python. The return of the episode was $-560.49$.}
  \label{fig:ppo-python-validation-745}
  \end{minipage}
  \hspace{0.05\textwidth}
  \begin{minipage}[b]{0.475\textwidth}
  \centering
  \begin{tikzpicture}[scale=0.675]
    \definecolor{xcolor}{RGB}{0,70,135};
    \definecolor{ycolor}{RGB}{255,70,15};
    \definecolor{phidotcolor}{RGB}{0,90,0};
  
    \begin{axis}[
        grid=major,
        xlabel={Time Step / - ($\Delta t = \SI{0.05}{\second}$)},
        ymin=-9, ymax=9,
        ytick distance=2,
        legend pos=north east
    ]
      \addplot[color=xcolor] table [x="Time Step", y="x", col sep=semicolon] {DDPG_Python_Validation_Cycle_45.csv};
      \addlegendentry{$x$ / $\SI{}{\meter}$}
      \addplot[color=ycolor] table [x="Time Step", y="y", col sep=semicolon] {DDPG_Python_Validation_Cycle_45.csv};
      \addlegendentry{$y$ / $\SI{}{\meter}$}
      \addplot[color=phidotcolor] table [x="Time Step", y="Angular Velocity", col sep=semicolon] {DDPG_Python_Validation_Cycle_45.csv};
      \addlegendentry{$\dot{\phi}$ / $\SI{}{\radian\per\second}$}
    \end{axis}
  \end{tikzpicture}
  \caption{Best validation at cycle 45 of the LExCI DDPG training with Python. The return of the episode was $-348.05$.}
  \label{fig:ddpg-python-validation-45}
  \end{minipage}
\end{figure}

\begin{figure}[h!]
  \begin{minipage}[b]{0.475\textwidth}
  \centering
  \begin{tikzpicture}[scale=0.675]
    \definecolor{xcolor}{RGB}{0,70,135};
    \definecolor{ycolor}{RGB}{255,70,15};
    \definecolor{phidotcolor}{RGB}{0,90,0};
  
    \begin{axis}[
        grid=major,
        xlabel={Time Step / - ($\Delta t = \SI{0.05}{\second}$)},
        ymin=-9, ymax=9,
        ytick distance=2,
        legend pos=north east
    ]
      \addplot[color=xcolor] table [x="Time Step", y="x", col sep=semicolon] {PPO_Simulink_Validation_Cycle_710.csv};
      \addlegendentry{$x$ / $\SI{}{\meter}$}
      \addplot[color=ycolor] table [x="Time Step", y="y", col sep=semicolon] {PPO_Simulink_Validation_Cycle_710.csv};
      \addlegendentry{$y$ / $\SI{}{\meter}$}
      \addplot[color=phidotcolor] table [x="Time Step", y="Angular Velocity", col sep=semicolon] {PPO_Simulink_Validation_Cycle_710.csv};
      \addlegendentry{$\dot{\phi}$ / $\SI{}{\radian\per\second}$}
    \end{axis}
  \end{tikzpicture}
  \caption{Best validation at cycle 710 of the LExCI PPO training with Simulink. The return of the episode was $-397.86$.}
  \label{fig:ppo-simulink-validation-745}
  \end{minipage}
  \hspace{0.05\textwidth}
  \begin{minipage}[b]{0.475\textwidth}
  \centering
  \begin{tikzpicture}[scale=0.675]
    \definecolor{xcolor}{RGB}{0,70,135};
    \definecolor{ycolor}{RGB}{255,70,15};
    \definecolor{phidotcolor}{RGB}{0,90,0};
  
    \begin{axis}[
        grid=major,
        xlabel={Time Step / - ($\Delta t = \SI{0.05}{\second}$)},
        ymin=-9, ymax=9,
        ytick distance=2,
        legend pos=north east
    ]
      \addplot[color=xcolor] table [x="Time Step", y="x", col sep=semicolon] {DDPG_Simulink_Validation_Cycle_45.csv};
      \addlegendentry{$x$ / $\SI{}{\meter}$}
      \addplot[color=ycolor] table [x="Time Step", y="y", col sep=semicolon] {DDPG_Simulink_Validation_Cycle_45.csv};
      \addlegendentry{$y$ / $\SI{}{\meter}$}
      \addplot[color=phidotcolor] table [x="Time Step", y="Angular Velocity", col sep=semicolon] {DDPG_Simulink_Validation_Cycle_45.csv};
      \addlegendentry{$\dot{\phi}$ / $\SI{}{\radian\per\second}$}
    \end{axis}
  \end{tikzpicture}
  \caption{Best validation at cycle 45 of the LExCI DDPG training with Simulink. The return of the episode was $-382.50$.}
  \label{fig:ddpg-simulink-validation-45}
  \end{minipage}
\end{figure}
\end{appendices}
\pagebreak



\end{document}